\documentclass[letterpaper, 10 pt, conference]{ieeeconf}  
\IEEEoverridecommandlockouts                               
\usepackage{graphicx}                       
\usepackage{graphics}                       
\usepackage{epsfig}                         
\usepackage{epstopdf}
\usepackage[tight,footnotesize]{subfigure}  
\usepackage{amssymb,amsmath}
\usepackage{mdwmath}
\usepackage{commath}                        
\usepackage{eqparbox}
\usepackage{amsmath,amsfonts,amssymb}

\newcommand{\eg}{{\em e.g.\ }}

\newcommand{\et}{{\em et al.}}

\newcommand{\beq}{\begin{equation}}
\newcommand{\eeq}{\end{equation}}
\newcommand{\bear}{\begin{eqnarray}}
\newcommand{\bears}{\begin{eqnarray*}}
\newcommand{\eear}{\end{eqnarray}}
\newcommand{\eears}{\end{eqnarray*}}
\newcommand{\bdm}{\begin{displaymath}}
\newcommand{\edm}{\end{displaymath}}
\newcommand{\lba}{\left[\begin{array}}
\newcommand{\ear}{\end{array}\right]}

\newcommand\given[1][]{\:#1\vert\:}    
\usepackage{stfloats}                       
\usepackage{hyperref}
\usepackage{cite}                           
\usepackage[T1]{fontenc} 
\usepackage{tabularx}
\usepackage[table]{xcolor}
\usepackage{longtable}
\usepackage{booktabs}       					
\usepackage{xcolor,colortbl}               		
\newcommand{\redz}{\textcolor[rgb]{ .153,  0,  .043}}
\newcommand{\reda}{\textcolor[rgb]{ .273,  0,  .043}}
\newcommand{\redb}{\textcolor[rgb]{ .393,  0,  .043}}
\newcommand{\redc}{\textcolor[rgb]{ .413,  0,  .043}}
\newcommand{\redd}{\textcolor[rgb]{ .543,  0,  .043}}
\newcommand{\rede}{\textcolor[rgb]{ .653,  0,  .043}}
\newcommand{\redf}{\textcolor[rgb]{ .773,  0,  .043}}
\title{\LARGE \bf Robot Introspection with Bayesian Nonparametric Vector Autoregressive Hidden Markov Models}
\author{Hongmin Wu, Hongbin Lin, Yisheng Guan, Kensuke Harada and Juan Rojas$^{*}$ \\
}
\begin{document}
\maketitle
\thispagestyle{empty}
\pagestyle{empty}
\bstctlcite{IEEEexample:BSTcontrol}
\begin{abstract}
Robot introspection, as opposed to anomaly detection typical in process monitoring, helps a robot understand what it is doing at all times. A robot should be able to identify its actions not only when failure or novelty occurs, but also as it executes any number of sub-tasks. As robots continue their quest of functioning in unstructured environments, it is imperative they understand what is it that they are actually doing to render them more robust.
This work investigates the modeling ability of Bayesian nonparametric techniques on Markov Switching Process to learn complex dynamics typical in robot contact tasks. We compare the Markov switching process, together with different observation model and the dimensionality of sensor data.
The work was tested in a snap assembly task characterized by high elastic forces. The task consists of an insertion subtask with very complex dynamics.
Our approach showed a stronger ability to generalize and was able to better model the subtask with complex dynamics in a computationally efficient way. The modeling technique is also used to learn a growing library of robot skills, one that when integrated with low-level control allows for robot online decision making. Supplemental info can be found at \cite{2017INIPS-Rojas-supplementalURL}.
\end{abstract}
\section{Introduction}
Robot introspection, as opposed to novelty detection typical in process monitoring, helps robots understand what it is they are doing and how they are doing it. Robots should identify actions both in nominal and anomalous situations. As robots continue their quest of functioning in unstructured environments, it is imperative they understand the "what" and the "how" of their actions in accurate, robust, and computationally tractable ways. Thus empowering them to improve their online decision making ability and better negotiate the unexpected.

Related to introspection is the area of high level state estimation or process monitoring. The vast majority of systems developed to model robot behaviors are parametric. This implies that system designers need to make strong assumptions about the underlying model including the number of hidden latent states (from now on referred to as the model complexity or simply complexity) and parameter values. The quality of the model depends on such parameters: if the complexity is too large or too small, the accuracy of the representation lessens and, the robot's insight is weakened. Furthermore, when the manipulation task is composed of actions that lead to highly dynamic effort (wrench or joint torque) signatures, such signals are much harder to model than those in which light contact or unperturbed motions take place. Markov models with their assumption on conditionally independent observations often cannot properly model such dynamics.

This work investigates the modeling ability of Bayesian nonparametric techniques on Markov Switching Process to learn complex dynamics typical in robot contact tasks. We compare the Markov switching process, together with different observation model and the dimensionality of sensor data. A stochastic prior: the sticky Hierarchical Dirichlet Process (sHDP)-is used to derive the model complexity (which represent primitive manipulation actions), while the Markov Switching Process known as the switching vector autoregressive (AR-HMM) is used to model complex dynamical phenomena encoded in the wrench signature of a robot performing contact tasks. We hypothesize the approach will lead to good generalizations that discover and model the underlying contact states in a computationally efficient manner. While others have used Bayesian nonparametric HMMs to model contact tasks, it seems no one has yet used the more expressive sHDP-VAR-HMM to model robot contact states and seek to endow both nominal and anomalous introspection to the robot.

The introspection system runs in parallel to a manipulation behavior that is encoded through a Finite State Machine (FSM). The FSM also receives feedback from the process monitor. A Bayesian non-parametric prior, conjugate to the categorical distribution, is used to learn the mode complexity and transition distribution of the latent state portion of the model for a given robot skill. With respect to the vector autoregressive, the Bayesian approach uses a conjugate prior to the AR likelihood to learn the posterior distribution of a set of dynamical parameters. For each state in the FSM, inference is used to acquire models of the various subtasks in the task. From the set of trained models, an expected log-likelihood is derived. A classification threshold for anomalous (or unexpected events) detection is also derived from the trained model expected log-likelihood. Testing looks for both correct introspection (of the subtask at hand) and possible anomalous behavior. To do so, log-likelihoods of cumulative observations of the testing trial given separate sets of the trained models are produced. Correct introspection occurs when the log-likelihood of the skill at hand (as indexed by an FSM) attains the largest likelihood, otherwise the nominal skill identification is faulty. Simultaneously, we monitor whether or not test trial log-likelihood belonging to the current skill crosses below an anomalous threshold. If it does, an anomaly is said to have occurred.

\begin{figure}
	\centering
		\subfigure{\label{fig:hiro_snap}\includegraphics[width=3.0in,height=1.4in]{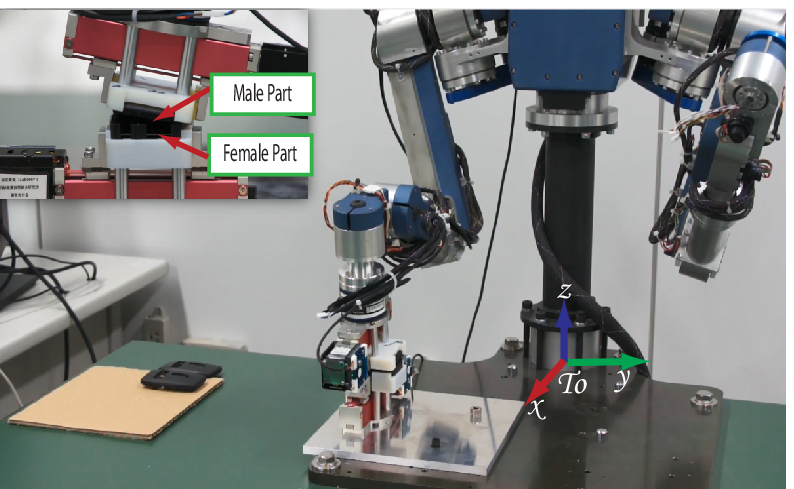}}
		\subfigure{\label{fig:sHDP-VAR-HMM}\includegraphics[width=3.0in,height=1.4in]{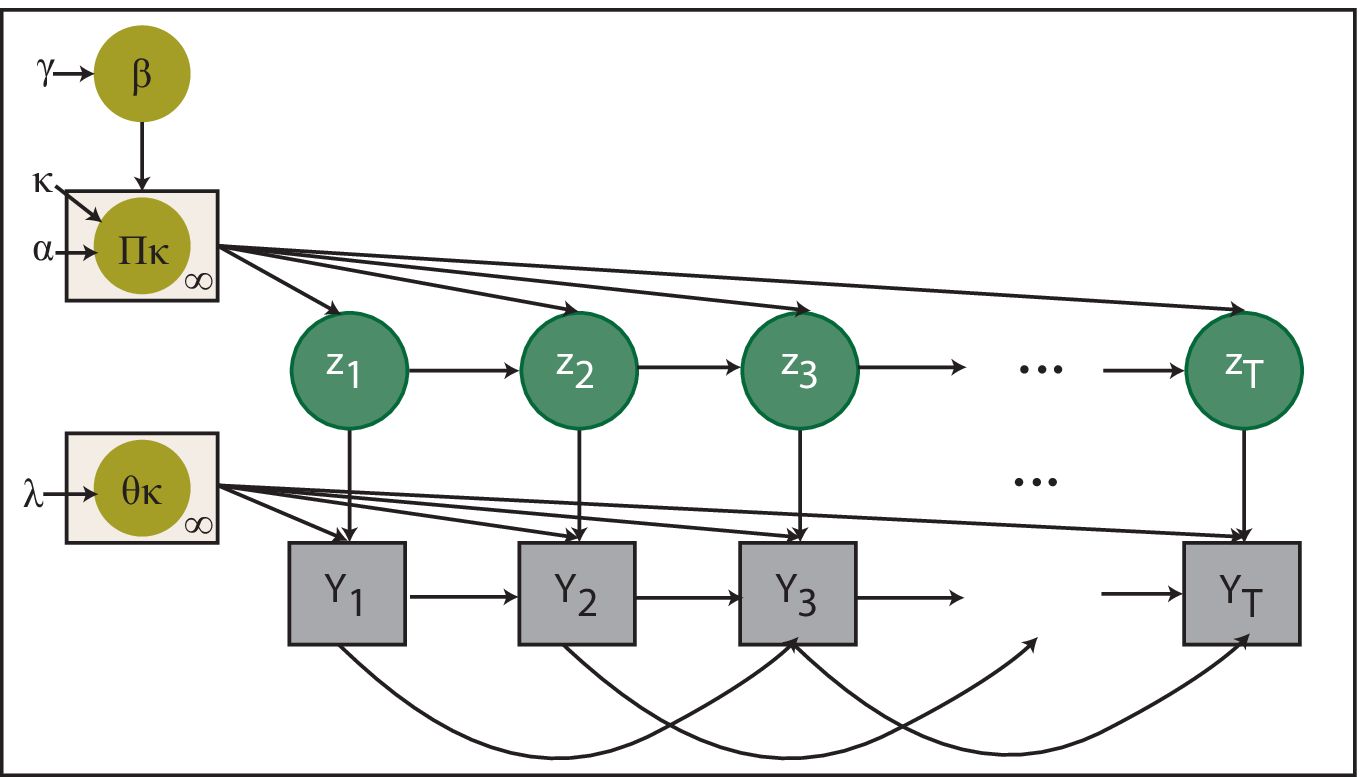}}
		\caption{On the top, the HIRO robot performs a snap assembly task. Both tasks are driven by a finite state machine\protect\footnotemark. On the bottom, a graphical representation of the sHDP-VAR-HMM used by the robot introspection system which runs in parallel to the low-level control to provide high-level semantic descriptions of the identified robot action. See Sec. \ref{sec:sHDP-VAR} for the mathematical treatment.}
        \label{fig:intro}
\end{figure}

A cantilever snap assembly task was used for demonstration (see Fig. \ref{fig:hiro_snap}. Both tasks were driven by FSMs in their nominal behavior. Anomalous cases in the snap assembly case occurred from time-to-time throughout experimentation.
sHDP-VAR-HMM results were compared with other nonparametric HMMs baseline.

The paper shows the nonparametric approach produced better and faster accuracy than baseline comparisons. Such modeling technique is then used to learn a library of robot skills that can be used for future introspection and integrated with low-level control for online decision making.
\footnote{Supplemental information, code, and data can be found at \cite{2017INIPS-Rojas-supplementalURL}.}

The remainder of the paper is structured as follows. Related works and limitations are introduced in Sec. \ref{sec:related_work}. The Bayesian nonparametric Markov switching process is described in Sec. \ref{sec:sHDP-VAR}. Experiments and results for a real-robot contact tasks is presented in Sec. \ref{sec:experiments}. Discussion of significance, strengths, and weaknesses are presented in Sec. \ref{sec:discussion}, and key points summarized in Sec. \ref{sec:conclusion}.
\footnotetext{http://www.ros.org/SMACH}
\section{Related Work}\label{sec:related_work}
Most of the work related to high level state estimation or process monitoring has used parametric models. Parametric models assume a fixed and prescribed model complexity (i.e. number of latent states in the model). Few works in process monitoring have studied nonparametric approaches that allow a prior with an unbounded number of potential model parameters to learn the model complexity based on the data. In doing so, one can expect better modeling and thus behavior identification.

Supervised learning problem, a common approach in process monitoring, maps input features to a fixed and prescribed number of manually annotated classes.
For contact tasks state estimation, Rojas \et, extract relative-change patterns classified through a small set of categories and aided by contextual information, and where increasingly abstract layers were used to estimate task behaviors \cite{2013IJMA-Rojas-TwrdsSnapSensing}. The framework used Bayesian models to provide a belief about its classification \cite{2012Humanoids-Rojas-pRCBHT} and SVMs to identify specific failure modes \cite{2017iros-rojas-onlinewrenchintrospection}. In \cite{2010CASE-Rodriguez-FailureDetAsmbly_ForceSigAnalysis}, Rodriguez \et, classified wrench data using SVM's to learn a decision rule between successful and failed assemblies offline. Later in \cite{2011IROS-Rodriguez-AbortRetry}, a probabilistic classification for outputs was produced through Relevance Vector Machines in combination with a Markov Chain model. In \cite{2015ICRA-Golz-TactileSensingLearnContactKnowledge}, Golz \et. used an SVM approach to distinguish regular human-robot interactions versus collisions, thus inferring the intention of the robot. Artificial Neural Networks (NN) have also been used to perform state estimation \cite{2008JMES-Althoefer-AutFailClassAsmblyThreadFastn}, \cite{2014JMES-Diryag-NN_PredictBotFail}. In \cite{2008JMES-Althoefer-AutFailClassAsmblyThreadFastn}, Althoefer \et, used a radial basis NN to monitor and estimate failure modes for threaded fastener assemblies. Most recently, Kappler \et, used a naive Bayes approach with a Gaussian model to incrementally model the current state of execution through kinesthetic demonstrations \cite{2015RSS-Kappler-DateDrivenOnlineDecisionMakingManipu}. The system tightly integrates a low-level control loop with a high-level state estimation loop to perform tractable online decision making. A supervised discriminative model is used to learn general failure but has to be taught, no global failure threshold exists at system deployment.

Others have opted to use a stochastic processes based on a Markov chain. Namely the Hidden Markov model or variants thereof. In \cite{1998IJRR-Hovland-HMM_ProcessMonitorAsmbly}, Hovland and McCarragher pioneered the use of HMMs to model contact events by observing wrench signatures. The contact state was identified among a set of discrete edge-surface configurations and provided a probability over a sequence of contacts. In this work, the model complexity remained fixed. Kroemer \et, learned to model the transitions amongst robot subtasks in manipulation problems \cite{2014ICRA-kroemer-LearnPredictPhasesManipHiddenStates},\cite{2015ICRA-Kroemer-TwrdsLearnHierSkillsMultiPhaseManip}. The work used a variant of the standard autoregressive hidden Markov model (AR-HMM) where actions, states, and robot skills were modeled. The key difference to the AR-HMM is the additional edge from the current state to the current skill. As a result of this edge, the transition between skills depends on the observed state. However, the results were still susceptible to the number of skills used and requires validation tests at the end to compute the correct number. Park \et, used an HMM with a multi-modal feature vector for execution monitoring and anomaly detection \cite{2016ICRA-Park-MultiModalMonitoringAnomalyDet_RobotManip}. They innovated an anomaly detection threshold whose value varied according to the task's progress.

Some have used nonparametric Bayesian approaches to Markov Switching Processes. In \cite{2012NIPS-DiLello-HDPHMM_AbnormalDetectionBotAsmbly}, \cite{2013IROS-DiLello-BayesianContFaultDetection}, DiLello \et, used the sticky hierarchical Dirichlet process prior to learn the model parameters of an HMM based on wrench signatures for an industrial robot alignment task. The work was used to learn specific failure modes and do online classification. The work did not explore more expressive Markov Switching Processes for modeling the robot signals nor did he focus on nominal identification of robot subtasks. In \cite{2015IJRR-Niekum-LrnGrnddFiniteStateReprUnstrucDems}, Niekum \et, used a beta-process prior on the vector autoregressive (BP-AR-HMM) to extract the mode complexity from the data. The BP-AR-HMM has the ability to discover and model primitives from multiply related time-series. Robot skill discrimination was reliable and consistent though this work only modeled robot pose in a primarily non-contact oriented task.
\section{The sHDP-VAR-HMM}\label{sec:sHDP-VAR}
Contact task modeling involves the interpretation of noisy wrench signals. Wrench noise is not well approximated by Gaussian noise and may contain latent patterns that stem from the knowledge of an expert task programmer or human demonstrator. Such patterns vary when the same task is executed by different agents. Our goal is, then, given roughly similar signals, to identify fundamental temporal patterns and model signal evolution to provide the robot with temporal introspection about its evolving high level state. If successful, a robot can use this information to reason about its next move: whether selecting the next skill or recovering from abnormal behaviors (internal or external). Few works explore the capability of characterizing both robot skills and failure simultaneously in the same architecture. Both are critical to help the robot have a more complete understanding of its actions. We are thus in need of powerful modeling techniques to characterize complex dynamical phenomena, capturing both spatial and temporal correlations for the data, while integrating beliefs about the system. The sHDP-VAR-HMM leverages the attributes of combinatorial stochastic processes (sHDP) with state-space descriptions of dynamical systems (AR-HMM) \cite{2010-Fox-BNP_LearningMarkovSwitchingProcesses} to achieve this goal. In this section, we will provide a brief introduction about HMMs, followed by Bayesian nonparametrics, and conclude with the integration of state-space models.

\subsection{Hidden Markov Models}\label{subsec:hmm}
HMMs have been a workhorse in pattern recognition able to encode probabilistic state-space models. The HMM is a stochastic process where a finite number of latent states have Markovian state transitions. Conditioned on the mode sequence, the model assumes (discrete or continuous) conditionally independent observations given the latent state. Let $z_t$ denote the latent state at time $t \in T$, and $\pi_j$ the mode-specific transition distribution for mode $j \in K$. Then, given the mode $z_t$, the observation $y_t$ is conditionally independent of observations and modes at other times. The generative process is described as:
\begin{eqnarray}
  z_t \given z_{t-1} \sim  \pi_{z_{t-1}}, \quad
  y_t \given z_t     \sim  F(\theta_{z_t})
  \label{eqtn:hmm_gen_model}
\end{eqnarray}
where, $F(\cdot)$ represents a family of distributions (e.g., the multinomial for discrete data, or the multivariate Gaussian for real-vector-valued data); $\theta$ are the mode specific emission parameters.

\subsection{Bayesian Nonparametric Priors}\label{subsec:bnp}
The HMM approach assumes a fixed model complexity-a restrictive assumption. In process monitoring, state estimation, or introspection, latent states can be set to represent robot primitives. It's clear that not all robot skills have the same number of primitives and that even the same skill might have variation when conducted under different conditions. To render the mode complexity flexible, priors are used on probability measures.

HMMs can also be represented though a set of transition probability measures $G_j$. Probability measures yield strictly positive probabilities and add to 1. Consider if instead of using a transition distribution on latent states, we use it across emission parameters $\theta_j \in \Theta$. Then, $G_j = \sum^K_{k=1} \pi_{jk} \delta_{\theta_k}$, where $\delta_{\theta_k}$ is the unit mass for mode $k$ at $\theta$. For emission parameter $\theta_j$, equate with time index $t-1$, such that:
\begin{eqnarray}
  \theta'_t \given \theta'_{t-1}   \sim  G_{j_{t-1}}, \quad
  y_t \given \theta'_t \sim  F(\theta'_t)
  \label{eqtn:probability_measures}
\end{eqnarray}
So, given an $\theta_j$, different probability weights are assigned to possible successor candidates $\theta_k$. We can also assign a prior to the categorical probability measure $G_j$. The Dirichlet distribution is a natural selection due to conjugacy. Thus, the transition probabilities $\pi_j=[ \pi_{j1} \cdots \pi_{jK} ]$ are independent draws from a K-dimensional Dirichlet distribution: $\pi_j \sim Dir(\alpha_1, ...,\alpha_K),\mbox{ where, }j=1,...,K.$ And the sum $\sum_k \pi_{jk}=1$ as intended. The \textbf{Dirichlet process} (DP) was used as the base measured instead of the Dirichlet distribution. The DP is a distribution over countably infinite probability measures $G_0:\Theta \Rightarrow R^+$, where $G_0(\theta)\geq=0$ and $\int_\Theta G_0(\theta)d\theta=1$ \cite{2012MIT-Murphy-ML_ProbPerspective}. The DP has a joint Dirichlet distribution $(G_0(\theta_j),...,G_0(\theta_K)) \sim Dir(\gamma H(\theta_j),...,\gamma H(\theta_K))$ and we summarize the probability measure as $G_0 \sim DP(\gamma,H)$ as:
\begin{eqnarray}
  G_0 = \sum_{k-1}^\infty \beta_k \delta_{\theta_k} \quad \theta_k \sim H,
  \label{eqtn:DP_prob_measuers}
\end{eqnarray}
where $\gamma$ is the concentration parameter and $H$ is the base measure over parameter space $\Theta$. The weights $\beta_k$ are sampled via a stick-breaking construction: $\beta_k=\nu_k \Pi_{l=1}^{k-1}(1-\nu_l)$, where $\nu_k \sim Beta(1,\gamma)$. For succinctness, the stick-breaking process is defined as: $\beta \sim GEM(\gamma)$. The DP is used to define a prior on the set of HMM transition probability measures $G_j$. However, if each transition measure $G_j$ is an independent draw from $DP(\gamma,H)$, where $H$ is continuous, like a Gaussian distribution, transition measures lead to non-overlapping support. This means that previously seen modes (robotic primitives) cannot be selected again. To deal with this limitation, a \textbf{Hierarchical Dirichlet Process} (HDP) is used. The latter constructs transition measures $G_j$ \textit{on the same support points} $(\theta_1,\theta_2,...\theta_K)$. This can be done when $G_j$ is only a variation on a \textit{global} discrete measure $G_0$, such that:
\begin{eqnarray}
	G_0=\sum_{k-1}^\infty \beta_k \delta_{\theta_k}, 	& \beta \given \gamma \sim GEM(\gamma) \mbox{, } \theta_k \given G_0 \sim G_0 	\nonumber \\
    G_j = \sum_{k-1}^\infty \pi_{jk} \delta_{\theta_k}, 	& \pi_j \given \alpha,\beta \sim DP(\alpha,\beta). & \nonumber
    \label{eqtn:hdp}
\end{eqnarray}
This HDP is used as a prior on the HMM. The implications are a mode complexity learned from the data and a sparse state representation. The $DP(\alpha,\beta)$ distribution encourages modes with similar transition distributions, but does not distinguish between self- and and cross-mode transitions. This is problematic for dynamical data. The HDP-HMM yields large posterior probabilities for mode sequences with unrealistically fast dynamics. Fox \et, \cite{2010-Fox-BNP_LearningMarkovSwitchingProcesses} introduced the sticky parameter into the HDP, yielding the \textbf{sHDP-HMM}. The latter increases the expected probability of self-transitions by an amount proportional to $\kappa$ and leads to posterior distributions with smoothly varying dynamics. Finally, priors are placed on the concentration parameters $\alpha$ and $\gamma$, and the sticky parameter $\kappa$. Latent state creation is influenced by $\alpha$ and $\gamma$, while self-transition probabilities are biased by $\kappa$. These priors allow to integrate specific system knowledge--an advantage compared to the expectation-maximization (EM) algorithm traditionally used in HMMs.
\subsection{The sHDP-AR-HMM}\label{subsec:AR-HMM}
The Autoregressive (AR) is a special case of Markov Jump Linear Switch systems, in which each dynamical mode is modeled by a linear dynamical system. That is, the observations are a noisy linear combination of a finite set of past observations with additive white noise. For an $r^{th}$ order vector, the process is VAR(r). Many complex dynamical phenomena cannot be adequately described by a single linear dynamical model; instead dynamics are approximated as switches between sets of linear systems in a stochastic process based on an underlying discrete-valued system mode. If the latent mode is a discrete-time Markov process, the model is a Markov Jump Linear system. The VAR(r) system can be considered an extension of the HMM in which each state is associated with a linear dynamical process. However, the VAR has simplifying assumptions that make it a practical choice in applications \cite{2010-Fox-BNP_LearningMarkovSwitchingProcesses}. The switching regime can be combined with the sHDP-HMM from Sec. \ref{subsec:bnp} to leverage the expressiveness of the VAR system with the ability of non-parametric priors to learn the mode complexity of the model. The VAR(r) process, with autoregressive order $r$ consists of a latent state $z_t \in R^n$ with linear dynamics observed through $y_t \in R^d$. The observations have mode specific coefficients and process noise as:
\begin{eqnarray}
	y_t=\sum_{i=1}^r A_i^{(z_t)} y_{t-i} + e_t(z_t), \quad e_t \sim \mathcal{N}(0,\Sigma).
    \label{eqtn:var}
\end{eqnarray}
We see a generative model for a time-series $\{y_1, y_2, ..., y_T\}$ of observed multi-modal data, a matrix of regression coefficients $\mathbf{A}^{(k)}=[A_1^{(k)} \cdots A_r^{(k)}] \in \mathbb{R}^{dx(d*r)}]$, and a measurement noise $\Sigma$, with a symmetric positive-definite covariance matrix. Given the observation data, we are interested in learning the ``$r^{th}$'' model order, for which we need to infer $ \{ \mathbf{A}^{(k)},\Sigma^{(k)} \}$. We leverage the Bayesian approach through the placement of conjugate priors on both parameters for posterior inference. As the mean and covariance are uncertain, the Matrix-Normal Inverse-Wishart (MNIW) serves as an appropriate prior on the multivariate AR distribution. The MNIW places a conditionally matrix-normal prior on $\mathbf{A}^{(k)}$ given $\Sigma$:
\begin{eqnarray}
	\mathbf{A}^{(k)} \given \Sigma \sim \mathcal{MN}(\mathbf{A};M,\Sigma,K),
    \label{eqtn:regression_coefficient_learning}
\end{eqnarray}
and an inverse-Wishart prior on $\Sigma$:
\begin{eqnarray}
	\Sigma \sim IW(\nu_0,S_0).
    \label{eqtn:covariance}
\end{eqnarray}
See \cite{2009PhD-Fox-BNP} Appendix F.1 for a detailed derivation of resulting posterior and definitions for variables $M$ and $K$ which are parameters that define the matrix-normal portion of the MNIW prior.
\section{Robot Introspection}\label{sec:introspection}
In our robot introspection system, we simultaneously detect nominal robot skills and possible anomalous unexpected events. Traditional robot behavior design uses FSMs to execute nominal robot tasks. FSMs partition the task into a sequence of skills (also known as sub-tasks or phases) and events that trigger a transition across skills. The skill sequence is learned from kinesthetic teaching or through an expert programmer and executed through a low-level controller (see our control basis controller \cite{2013IJMA-Rojas-TwrdsSnapSensing_A}. Transition thresholds are empirically devised under controlled conditions during the design-phase and usually focus on robot data such as lapsed-time, joint angle configurations, joint torques, or end-effector wrench change-point events. This setup is fragile, particularly in unstructured scenarios like human-robot collaboration, where many unexpected events may occur. Thus, there is a need for the robot both to learn new skills along with their identification and the detection of anomalous events.
\subsection{Model Training}\label{subsec:model_training}
We train our model on individual subtasks and capture their dynamics through an observation vector of $\tau_m$ end-effector wrench values and their first derivatives. The autoregressive model in the sHDP-AR-HMM captures spatio-temporal correlations in the observations. Blocked Gibbs sampling \cite{2009PhD-Fox-BNP} is used to learn the posterior distribution of the sHDP-AR-HMM along with mean values for the model parameters $\Pi$ of a given skill $s$, hence $\Pi_s=\{\pi,\textbf{A}\}$; that is, the transition matrix and the regressor coefficients.
\subsection{Nominal Skill Identification}\label{subsec:nominal_skills}
\begin{figure*}[bt]
	\centering		
		\includegraphics[width=0.8\linewidth]{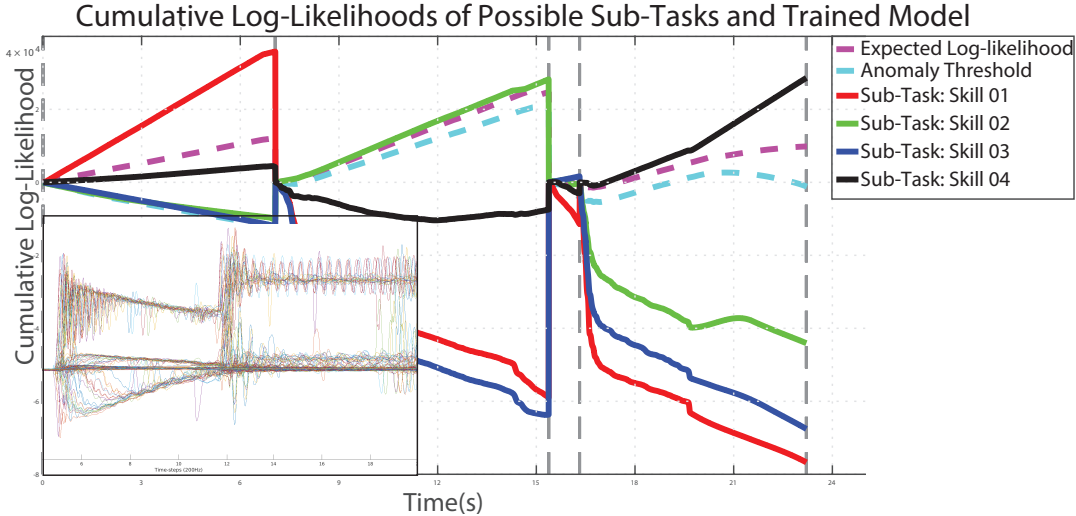}
		\caption{Cumulative log-likelihoods using test trial observations conditioned on all available trained skill model parameters $log P(y_{s_1:s_t} \given \Pi)_s^S$. Wrench signals used for training used in lower-left corner. Thresholds used: (i) Expected log-likelihood in cyan dashed line, (ii) Anomaly detection set by $\mu -k*\sigma$ from the expected log. If the current likelihood dips under threshold an anomaly is reported.}
        \label{fig:likelihoods}
\end{figure*}
Given $S$ trained models for $M$ robot skills, k-fold cross validation is used along with the standard forward-backward algorithm to compute the \textit{expected cumulative likelihood} of a sequence of observations $\mathbb{E} \left[ log \mbox{ } P(Y \given \Pi_s) \right]$ for each trained model $s \in S$ as illustrated in Fig. \ref{fig:likelihoods}. Given a test trial $r$, the cumulative log-likelihood is computed for test trial observations conditioned on all available trained skill model parameters $log \mbox{ } P(y_{r_1:r_t} \given \Pi)_s^S$ at a rate of 200Hz. The process is repeated when a new skill $m$ is started. Given the position in the FSM $s_c$, we can index the correct log-likelihood $\mathbb{I}(\Pi_s=s_c)$ and see if its probability density of the test trial given the correct model is greater than the rest:
\begin{eqnarray}
	log \mbox{ } P(y_{r_1:r_t} \given \Pi_{correct}) > log \mbox{ } P(y_{r_1:r_t} \given \Pi_s), \\ \nonumber
	\quad \forall s(s \in S \land s \neq s_c).
    \label{eqtn:state_classification_condition}
\end{eqnarray}
If so, the identification is deemed correct, and the time required to achieve the correct classification recorded. At the end of the cross-validation period, a classification accuracy matrix is derived as well as the mean time threshold value (see results in Sec \ref{sec:experiments}). In Fig. \ref{fig:likelihoods}, one can appreciate the log-likelihood curves for a snap assembly task consisting of 4 sub-tasks: a guarded approach, an alignment, an insertion, and mating. Note how for each of the subtasks, the log-likelihood corresponding to the current FSM index, has a greater value than the rest.
\subsection{Anomaly Identification}\label{subsec:anomaly_identification}
General anomaly detection vs specific failure mode identification is considered a better choice. Specific failure mode classification has many disadvantages: an infinite number of failure modes (even a small representative subset requires much work); intentional failure, which is difficult to execute puts the robot at risk. Learning failures as you go is viable \cite{2015RSS-Kappler-DateDrivenOnlineDecisionMakingManipu}, but we deem that general anomaly detection complemented with recovery strategies based on additional sensory information is easier to generalize and more robust. Some have even began leveraging human sense of failure to guide robots \cite{2017Science-Salazar-Correcting_Robot_Mistakes_in_Real_Time_using_EEG_Signals}.

In our work, anomaly detection is continuously monitored for as in \cite{2016ICRA-Park-MultiModalMonitoringAnomalyDet_RobotManip}. It is based on the notion that in nominal tasks, the cumulative likelihood has similar patterns across trials of the same robot skill. Thus, the expected cumulative log-likelihood $L$ derived in training can be used to implement an anomaly threshold $F$. In Fig. \ref{fig:likelihoods}, the dashed cyan curve represents $L$ generated from training data for each skill. For each time step in an indexed skill $s_c$, the anomaly threshold is set to $F_{s_c}=\mu(L)-k*\sigma(L)$, where $k$ is a real-valued constant that is multiplied by the standard deviation to change the threshold. We are only interested in the lower (negative) bound. If the cumulative likelihood crosses the threshold, an anomaly is flagged: $\mbox{if } log \mbox{ } P(y_{r_1:r_t}  \given \Pi_{correct}) < F_{s_c} \mbox{: anomaly, else nominal}$.
\section{Experiments and Results}\label{sec:experiments}
Hiro, a 6 DoF dual-arm robot with electric actuators and a JR3 6DoF force-torque (FT) sensor attached on the wrist is used to perform a snap assembly task of camera parts. A custom end-effector holds the male part, while the female is fixed to a table in front of the robot (see Fig. \ref{fig:hiro_snap}). OpenHRP \cite{2004IJRR:Kaneheiro:OpenHRP} executes the FSM and modular hybrid pose-force-torque controllers \cite{2013IJMA-Rojas-TwrdsSnapSensing} execute the skills. Four nominal skills are connected by the FSM: (i) a guarded approach, (ii) an alignment procedure, (iii) a snap insertion with high elastic forces, and (iv) a mating procedure. Unexpected events occur during initial parts' contact, (\eg wrong parts localization) or during the insertion stage where wedging is possible.
The tool center point (TCP) was placed at the point where the male parts contacts the female. The world reference frame was located at the manipulator's base. The TCP position and orientation were determined with reference to the world coordinate frame $To$. The force and torque reference frames were determined with respect to the wrist's reference frame.

Results for skill identification and anomaly detection are report independently for clarity. 44 real-robot nominal trials and 16 anomalous trials were conducted.
Leave-one-out cross-validation was used to train both the subtask and anomaly identification. In the case of anomaly detection, the trials within the test folds for skill and anomaly identification were randomly intermixed.
\begin{table*}[bt]
\centering
\caption{Classification accuracy table for different skills in the snap assembly task. The table compares different HMM algorithms (parametric and non-parametric) across different sensor modalities for the observation vector. A computation of how fast a decision can be made is registered as a percentage of the execution length of a skill.}
\label{tab:confusion_matrix}
\begin{tabular}{lllll}
Classification Accuracy                                     & S\_01   & S\_02  & S\_03   & S\_04  \\
\midrule
sHDP-HMM (Wrench,Wrench 1st Derivative)                     & 80\%    & 100\%  & \reda{0\%}     & 100\%  \\
sHDP-HMM (Wrench,Wrench 1st Derivative,Pose)        & 85\%    & 90\%   & \redb{60\%}    & 100\%  \\
sHDP-VAR(1)-HMM (Wrench,Wrench 1st Derivative)              & 100\%   & 100\%  & \redc{70\%}    & 100\%  \\
sHDP-VAR(1)-HMM (Wrench,Wrench 1st Derivative,Pose) & 100\%   & 100\%  & \redd{90\%}    & 100\%  \\
sHDP-VAR(2)-HMM (Wrench,Wrench 1st Derivative)              & 100\%   & 100\%  & \rede{80\%}    & 100\%  \\
sHDP-VAR(2)-HMM (Wrench,Wrench 1st Derivative,Pose) & 100\%   & 100\%  & \redf{95\%}    & 100\%  \\
\midrule
Decision Time (lower better)                                &         &        &         &        \\
\midrule
sHDP-HMM (Wrench,Wrench 1st Derivative)                     & 27.53\% & 1.20\% & \redz{98.60\%} & 8.13\% \\
sHDP-HMM (Wrench,Wrench 1st Derivative,Pose)        & 18.42\% & 3.26\% & \reda{10.70\%} & 7.26\% \\
sHDP-VAR(1)-HMM (Wrench,Wrench 1st Derivative)              & 1.42\%  & 1.20\% & \redb{10.60\%} & 1.34\% \\
sHDP-VAR(1)-HMM (Wrench,Wrench 1st Derivative,Pose) & 16.73\% & 1.20\% & \redc{10.60\%} & 1.34\% \\
sHDP-VAR(2)-HMM (Wrench,Wrench 1st Derivative)              & 1.42\%  & 1.20\% & \redd{10.70\%} & 1.48\% \\
sHDP-VAR(2)-HMM (Wrench,Wrench 1st Derivative,Pose) & 1.42\%  & 1.20\% & \rede{10.64\%} & 1.53\%
\end{tabular}
\end{table*}

A 12-dimensional observation vector captured wrench and wrench $1^{st}$ derivatives. We first test an HDP-VAR(1)-HMM with a MNIW prior with conservative values similar to those in \cite{2009PhD-Fox-BNP}. Parameters mean matrix $M$ and $K$ are set such that the mass of the matrix-normal distribution is centered around stable dynamic matrices while allowing variability in the matrix values (see Sec. \ref{subsec:AR-HMM} for details). $M=\mathbf{0}$ and $K = 10*I_{m}$. The inverse-Wishart portion of the prior is given by $\nu_0=m+2$ DoF (the smallest integer setting to maintain a proper prior). The scale matrix $S_0=0.75*$empirical covariance (of the data set). The setting is motivated by the fact that the covariance is computed from polling all of the data and it tends to overestimate mode-specific covariances. A value slightly less than 1 of the constant in the scale matrix mitigates the overestimation. Also, setting the prior from the data can move the distribution mass to reasonable parameter space values. A Gamma(a, b) prior is set on sHDP concentration parameters $\alpha+\kappa$ and $\gamma$. A $Beta(c,d)$ prior is set on the self-transition parameter $\rho$. We choose a weekly informative setting and choose: $a = 1, b = 0.01, c =10, d =1$. The Gibbs sampling parameters truncation level and maximum iterations are set to 20 and 500 respectively.
\subsection{Results}
Skill and anomaly identification results were across class of undirected graph algorithms, namely: an sHDP-HMM with different sets of sensory signals, and an sHDP-VAR-HMM with two regressive orders and sets of sensory signals. Table \ref{tab:confusion_matrix} presents a classification accuracy matrix, where identification accuracy per robot skill is presented and compared. The average time that for computing correct classifications is also reported as a percentage of the total duration of a skill. Low percentages imply quick classification, while large ones imply slow decisions. For anomaly classification, ROC curves were used to measure the discriminative ability of the system.
\section{Discussion}\label{sec:discussion}
The results show that the sHDP-AR-HMM with the greatest regressive order and additional sensory signals had the best identification accuracy than the rest of the methods, and close to the fastest speed. The algorithm was specially effective in discerning the Insertion subtask of the snap assembly task which, is the most challenging to model due to complex dynamics. The multi-modal sHDP-AR(2)-HMM outperforms the 1st order algorithm by 5\% and by 15\% and 25\% corresponding methods that only used wrench values and first derivatives of wrench values but not pose. The approach also classified nominal states $38\%$ faster than the next best model across all subtasks. On a different note, receiver operating characteristic (ROC) curves show that our model has better discrimination with less false positives than the rest of the models. The sHDP-AR-HMM thus makes stronger generalizations and is a more powerful modeling system than the sHDP-HMM and HMM.

\begin{figure}[tb]
      \centering
          \includegraphics[scale=0.65]{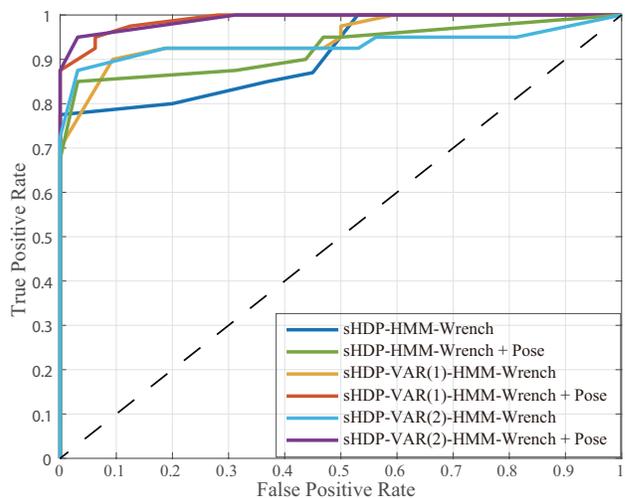}
      \caption{ROC Curve for anomaly detection for different graphical models. The ROC curve for the sHDP-VAR-HMM methods has the least area over the curve, indicating that overall it has a lower false-positive rate in detecting anomalies than the other methods.}
      \label{fig:anomaly_log-likelihoods}
\end{figure}

Compared to the SVM results in \cite{2017iros-rojas-onlinewrenchintrospection}, the SVM tends to do better, on the other hand, this system allows for the simultaneous checking of nominal and anomalous skills, not so in \cite{2017iros-rojas-onlinewrenchintrospection}, where a two stage classifier is used. Having said that, the SVM probabilistic method is able to provide a confidence parameter beyond accuracy classification, something that is not readily available in HMMs, though condition numbers seem possible but were not attempted in this work. We can only make a slight comparison with the work of \cite{2013IROS-DiLello-BayesianContFaultDetection} as in his work he only estimates one nominal skill (and 4 anomalous ones) with corresponding accuracies of $(97.5,90,90,90,80)\%$, with an overall average of 89.5\%. The task's nature is that of alignment and do not experience as complex dynamics as in an insertion task. In our work, our overall accuracy using multi-modal signals and the 2nd order regressive was 98.75\%.
In \cite{2013MICCAI-Ahmidi-StringMotifDescrToolMotion_SkillGestures}, Ahmidi \et tried to perform surgeon task monitoring in surgeries using the Da Vinci machine. Using Frenet-Frames they find ways to segment and encode position data and then classify it through graphical models. For their DCC19 experiments on data set DS-I, with known state boundaries and for spatial quantization, their average behavior classification rate was: 82.16\%. For their second data set DS-II, for spatial DCC19,their average classification results across k-fold validation and one-user out validation (for all their skill sets) was 75.22\%. This suggests that the results attained in this work
\section{Conclusion}\label{sec:conclusion}
In this work, the use of a non-parametric Markov switching process, the sHDP-AR-HMM, was used to perform robot introspection. The introspection system was able to identify nominal executing robot skills as well as the occurrence of anomalous activity. The combination of the VAR-HMM Markov-switching process, along with the nonparametric Bayesian prior used to learn posterior models, resulted in systems capable to  model complex switching dynamical phenomena with better and faster classification rates.
\section{Acknowledgments}
The work in this paper is supported by the NSFC-Guangdong Joint Fund (Grant No. U1401240), the Natural Science Foundation of Guangdong Province (Grant No. 2015A030308011), the National Natural Science Foundation of China (Grant No. 51605096), the State International Science and Technology Cooperation Special Items (Grant No. 2015DFA11700), and the Frontier and Key Technology Innovation Special Funds of Guangdong Province (Grant No. 2014B090919002, 2016B0911006, 2015B010917003, 2017B050506008).
\bibliographystyle{IEEEtran}
\bibliography{IEEEabrv,Xbib}

\begin{thebibliography}{10}
\providecommand{\url}[1]{#1}
\csname url@samestyle\endcsname
\providecommand{\newblock}{\relax}
\providecommand{\bibinfo}[2]{#2}
\providecommand{\BIBentrySTDinterwordspacing}{\spaceskip=0pt\relax}
\providecommand{\BIBentryALTinterwordstretchfactor}{4}
\providecommand{\BIBentryALTinterwordspacing}{\spaceskip=\fontdimen2\font plus
\BIBentryALTinterwordstretchfactor\fontdimen3\font minus
  \fontdimen4\font\relax}
\providecommand{\BIBforeignlanguage}[2]{{%
\expandafter\ifx\csname l@#1\endcsname\relax
\typeout{** WARNING: IEEEtran.bst: No hyphenation pattern has been}%
\typeout{** loaded for the language `#1'. Using the pattern for}%
\typeout{** the default language instead.}%
\else
\language=\csname l@#1\endcsname
\fi
#2}}
\providecommand{\BIBdecl}{\relax}
\BIBdecl

\bibitem{2017INIPS-Rojas-supplementalURL}
\BIBentryALTinterwordspacing
J.~Rojas, ``Supplement to robot introspection with bayesian nonparametric
  vector autoregressive hidden markov models,'' 2017. [Online]. Available:
  \url{http://www.juanrojas.net/shdp-var-hmm/}
\BIBentrySTDinterwordspacing

\bibitem{2013IJMA-Rojas-TwrdsSnapSensing}
J.~Rojas, K.~Harada, H.~Onda, N.~Yamanobe, E.~Yoshida, K.~Nagata, and Y.~Kawai,
  ``Towards snap sensing,'' \emph{International Journal of Mechatronics and
  Automation}, vol.~3, no.~2, pp. 69--93, 2013.

\bibitem{2012Humanoids-Rojas-pRCBHT}
J.~Rojas, K.~Harada, H.~Onda, N.~Yamanobe, E.~Yoshida, K.~Nagata, and Y.~Kawai,
  ``Probabilistic state verification for snap assemblies using the
  relative-change-based hierarchical taxonomy,'' in \emph{2012 12th IEEE-RAS
  International Conference on Humanoid Robots (Humanoids 2012)}.\hskip 1em plus
  0.5em minus 0.4em\relax IEEE, Nov 2012, pp. 96--103.

\bibitem{2017iros-rojas-onlinewrenchintrospection}
J.~Rojas, S.~Luo, D.~Zhu, Y.~Du, H.~Lin, Z.~Huang, W.~Kuang, and K.~Harada,
  ``Online robot introspection via wrench-based action grammars. *to appear in
  iros 2017.'' \emph{preprint arXiv:1702.08695}, 2017.

\bibitem{2010CASE-Rodriguez-FailureDetAsmbly_ForceSigAnalysis}
A.~Rodriguez, D.~Bourne, M.~Mason, G.~F. Rossano, and J.~Wang, ``Failure
  detection in assembly: Force signature analysis,'' in \emph{Automation
  Science and Engineering (CASE), 2010 IEEE Conference on}.\hskip 1em plus
  0.5em minus 0.4em\relax IEEE, 2010, pp. 210--215.

\bibitem{2011IROS-Rodriguez-AbortRetry}
A.~Rodriguez, M.~T. Mason, S.~S. Srinivasa, M.~Bernstein, and A.~Zirbel,
  ``Abort and retry in grasping,'' in \emph{Intelligent Robots and Systems
  (IROS), 2011 IEEE/RSJ International Conference on}.\hskip 1em plus 0.5em
  minus 0.4em\relax IEEE, 2011, pp. 1804--1810.

\bibitem{2015ICRA-Golz-TactileSensingLearnContactKnowledge}
S.~Golz, C.~Osendorfer, and S.~Haddadin, ``Using tactile sensation for learning
  contact knowledge: Discriminate collision from physical interaction,'' in
  \emph{Robotics and Automation (ICRA), 2015 IEEE International Conference
  on}.\hskip 1em plus 0.5em minus 0.4em\relax IEEE, 2015, pp. 3788--3794.

\bibitem{2008JMES-Althoefer-AutFailClassAsmblyThreadFastn}
K.~Althoefer, B.~Lara, Y.~Zweiri, and L.~Seneviratne, ``Automated failure
  classification for assembly with self-tapping threaded fastenings using
  artificial neural networks,'' \emph{Proc. of the Inst. of Mech. Engineers,
  Part C: Journal of Mechanical Engineering Science}, vol. 222, no.~6, pp.
  1081--1095, 2008.

\bibitem{2014JMES-Diryag-NN_PredictBotFail}
A.~Diryag, M.~Miti{\'c}, and Z.~Miljkovi{\'c}, ``Neural networks for prediction
  of robot failures,'' \emph{Proceedings of the Institution of Mechanical
  Engineers, Part C: Journal of Mechanical Engineering Science}, vol. 228,
  no.~8, pp. 1444--1458, 2014.

\bibitem{2015RSS-Kappler-DateDrivenOnlineDecisionMakingManipu}
D.~Kappler, P.~Pastor, M.~Kalakrishnan, M.~Wuthrich, and S.~Schaal,
  ``Data-driven online decision making for autonomous manipulation,'' in
  \emph{Proceedings of Robotics: Science and Systems}, Rome, Italy, 2015.

\bibitem{1998IJRR-Hovland-HMM_ProcessMonitorAsmbly}
G.~E. Hovland and B.~J. McCarragher, ``Hidden markov models as a process
  monitor in robotic assembly,'' \emph{The International Journal of Robotics
  Research}, vol.~17, no.~2, pp. 153--168, 1998.

\bibitem{2014ICRA-kroemer-LearnPredictPhasesManipHiddenStates}
O.~Kroemer, H.~Van~Hoof, G.~Neumann, and J.~Peters, ``Learning to predict
  phases of manipulation tasks as hidden states,'' in \emph{Robotics and
  Automation (ICRA), 2014 IEEE International Conference on}.\hskip 1em plus
  0.5em minus 0.4em\relax IEEE, 2014, pp. 4009--4014.

\bibitem{2015ICRA-Kroemer-TwrdsLearnHierSkillsMultiPhaseManip}
O.~Kroemer, C.~Daniel, G.~Neumann, H.~van Hoof, and J.~Peters, ``Towards
  learning hierarchical skills for multi-phase manipulation tasks,'' in
  \emph{International Conference on Robotics and Automation (ICRA)}, 2015.

\bibitem{2016ICRA-Park-MultiModalMonitoringAnomalyDet_RobotManip}
D.~Park, Z.~Erickson, T.~Bhattacharjee, and C.~C. Kemp, ``Multimodal execution
  monitoring for anomaly detection during robot manipulation,'' in \emph{2016
  IEEE International Conference on Robotics and Automation (ICRA)}, May 2016,
  pp. 407--414.

\bibitem{2012NIPS-DiLello-HDPHMM_AbnormalDetectionBotAsmbly}
H.~B. Enrico~DiLello, Tinne De~Laet, ``Hdp-hmm for abnormality detection in
  robotic assembly,'' in \emph{NIPS Workshop on Bayesian Nonparametric Models
  for Reliable Planning and Decision-Making under Uncertainty}, Dec. 2012.

\bibitem{2013IROS-DiLello-BayesianContFaultDetection}
E.~Di~Lello, M.~Klotzbucher, T.~De~Laet, and H.~Bruyninckx, ``Bayesian
  time-series models for continuous fault detection and recognition in
  industrial robotic tasks,'' in \emph{Intelligent Robots and Systems (IROS),
  2013 IEEE/RSJ International Conference on}.\hskip 1em plus 0.5em minus
  0.4em\relax IEEE, 2013, pp. 5827--5833.

\bibitem{2015IJRR-Niekum-LrnGrnddFiniteStateReprUnstrucDems}
S.~Niekum, S.~Osentoski, G.~Konidaris, S.~Chitta, B.~Marthi, and A.~G. Barto,
  ``Learning grounded finite-state representations from unstructured
  demonstrations,'' \emph{The International Journal of Robotics Research},
  vol.~34, no.~2, pp. 131--157, 2015.

\bibitem{2010-Fox-BNP_LearningMarkovSwitchingProcesses}
E.~B. Fox, E.~B. Sudderth, M.~I. Jordan, and A.~S. Willsky, ``Bayesian
  nonparametric methods for learning markov switching processes,'' \emph{Signal
  Processing Magazine, IEEE}, vol.~27, no.~6, pp. 43--54, 2010.

\bibitem{2012MIT-Murphy-ML_ProbPerspective}
K.~P. Murphy, \emph{Machine Learning : a probabilistic perspective},
  T.~Dietterich, Ed.\hskip 1em plus 0.5em minus 0.4em\relax MIT Press, 2012.

\bibitem{2009PhD-Fox-BNP}
E.~B. Fox, ``Bayesian nonparametric learning of complex dynamical phenomena,''
  Ph.D. dissertation, Massachusetts Institute of Technology, 2009.

\bibitem{2013IJMA-Rojas-TwrdsSnapSensing_A}
omitted, ``for review,'' \emph{omitted}, vol.~00, no.~2, pp. 00--00, 9999.

\bibitem{2017Science-Salazar-Correcting_Robot_Mistakes_in_Real_Time_using_EEG_Signals}
A.~F. Salazar-Gomez, J.~DelPreto, S.~Gil, F.~H. Guenther, and D.~Rus,
  ``Correcting robot mistakes in real time using eeg signals,'' \emph{ICRA.
  IEEE}, 2017.

\bibitem{2004IJRR:Kaneheiro:OpenHRP}
F.~Kanehiro, H.~Hirukawana, and S.~Kajita, ``Openhrp: Open architecture
  humanoid robotics platform,'' \emph{Intl. J. of Robotics Res.}, vol.~23,
  no.~2, pp. 155--165, 2004.

\bibitem{2013MICCAI-Ahmidi-StringMotifDescrToolMotion_SkillGestures}
N.~Ahmidi, Y.~Gao, B.~B{\'e}jar, S.~S. Vedula, S.~Khudanpur, R.~Vidal, and
  G.~D. Hager, ``String motif-based description of tool motion for detecting
  skill and gestures in robotic surgery,'' in \emph{Medical Image Computing and
  Computer-Assisted Intervention--MICCAI 2013}.\hskip 1em plus 0.5em minus
  0.4em\relax Springer, 2013, pp. 26--33.

\end{thebibliography}

\end{document}